\pdfoutput=1

\documentclass[11pt]{article}
\usepackage{enumitem}
\usepackage{makecell}
\usepackage[final]{acl}
\usepackage{xcolor}
\usepackage{tabularx}
\usepackage{booktabs}
\usepackage{subcaption}
\usepackage{graphicx}
\usepackage[most]{tcolorbox}
\usepackage{times}
\usepackage{placeins}
\usepackage{latexsym}
\usepackage{lipsum}
\usepackage{amsmath}
\usepackage{booktabs}
\usepackage{multirow}
\usepackage{placeins} 
\usepackage{float}
\usepackage[T1]{fontenc}

\usepackage[utf8]{inputenc}

\usepackage{microtype}

\usepackage{inconsolata}

\usepackage{graphicx}

\title{From Perceptions to Decisions: Wildfire Evacuation Decision Prediction with Behavioral Theory-informed LLMs}

\author{ 
  \textbf{Ruxiao Chen}\textsuperscript{*1}, 
  \textbf{Chenguang Wang}\textsuperscript{1}\thanks{Equal Contribution.}, 
  \textbf{Yuran Sun}\textsuperscript{2}, 
  \textbf{Xilei Zhao}\textsuperscript{2}, 
   \textbf{Susu Xu}\textsuperscript{1}\\
  \textsuperscript{1}Johns Hopkins University \textsuperscript{2}University of Florida\\
  \{rchen117, cwang274, susuxu\}@jhu.edu \\
}

\begin{document}
\maketitle
\begin{abstract}
Evacuation decision prediction is critical for efficient and effective wildfire response by helping emergency management anticipate traffic congestion and bottlenecks, allocate resources, and minimize negative impacts.
Traditional statistical methods for evacuation decision prediction fail to capture the complex and diverse behavioral logic of different individuals. 
In this work, for the first time, we introduce \emph{FLARE}, short for facilitating LLM for advanced reasoning on wildfire evacuation decision prediction, a Large Language Model (LLM)-based framework that integrates behavioral theories and models to streamline the Chain-of-Thought (CoT) reasoning and subsequently integrate with memory-based Reinforcement Learning (RL) module to provide accurate evacuation decision prediction and understanding. 
Our proposed method addresses the limitations of using existing LLMs for evacuation behavioral predictions, such as limited survey data, mismatching with behavioral theory, conflicting individual preferences, implicit and complex mental states, and intractable mental state-behavior mapping. 
Experiments on three post-wildfire survey datasets show an average of 20.47\% performance improvement over traditional theory-informed behavioral models, with strong cross-event generalizability. 
Our complete code is publicly available at \url{https://github.com/SusuXu-s-Lab/FLARE}
\end{abstract}

\section{Introduction}
Wildfires are emerging as a significant natural hazard worldwide~\cite{jain2020review,zahura2024impact}.
In the January 2025 Southern California wildfires, more than 200,000 residents received evacuation orders to leave their homes~\cite{nbcnews_california_wildfires_2025}.
There is an urgent demand for emergency planners and policymakers to develop effective evacuation strategies to mitigate wildfire impacts~\cite{mockrin2018does, tapley2023reinforcement}. However, successful evacuations require a clear understanding of the human evacuation decision-making process and outcomes (i.e. whether individuals will follow the order to evacuate or stay) during these events to help policymakers improve evacuation order design, develop more efficient emergency response strategies, and build more resilient communities~\cite{collins2018evacuation, lovreglio2020calibrating, hong2020modeling, sun4953233investigating}. 
risks\cite{lovreglio2020calibrating,sun4953233investigating}.

Previous studies often construct evacuation choice models through a conceptual framework, Protective Action Decision Model (PADM)~\cite{strahan2019protective, lovreglio2019modelling, santana2021psychological, SUN2024106557}, 
which are designed to incorporate psychological factors, like individual risk perception and threat assessment, into the prediction process. Based on the PADM framework, past methods may employ various statistical models (e.g., logistic regression~\cite{forrister2024analyzing}, multinomial logistic regression~\cite{mccaffrey2018should}) to predict individual-evacuation decisions using socio-demographic information as inputs, trained on post-wildfire survey data. 
However, these traditional PADM-type statistical models lack reasoning capabilities to capture the diverse and complex logic underlying human decision-making due to limited data and restrictive modeling structure, even when the survey design is grounded in established behavioral theories. In addition, these statistical methods struggle to integrate qualitative descriptions, such as narrative accounts of wildfire dynamics or contextual details, which are critical for understanding evacuees' perceptions and the rationale behind their evacuation decisions.

To address these limitations, the recent emergence of Large Language Models (LLMs) provides exceptional reasoning capabilities to model and predict evacuation decision-making processes~\cite{huang2022towards, nguyen2024predicting, liu2024large, lee2024reasoning}. Compared to traditional statistical models, LLMs display theory of mind (ToM) capabilities and have the potential to bridge the information gap present in survey data by better approximating human decision-making logic.
LLMs also facilitate the integration of contextual information into the predictive process.

However, employing the existing LLM framework for evacuation decision-making modeling and prediction, with socio-demographic information as inputs, still faces four significant challenges:
(1) \textbf{Mismatching with behavioral theory}: 
Evacuation survey data size is often limited, for example, 334 valid examples for the 2021 Marshall wildfires~\cite{forrister2024analyzing}. LLMs tend to capture only partial reasoning patterns and overfit limited survey data, struggling to align with established behavioral theories~\cite{tjuatja2023llms, petrov2024limited, macmillan2024ir}. 

(2) \textbf{Conflicting preferences in aligning with human thought}:
Reinforcement Learning with Human Feedback (RLHF) offers a promising approach to aligning LLM reasoning with human thought~\cite{sun2023aligninglargemultimodalmodels, zhang2024personalizationlargelanguagemodels, xu2024surveyknowledgedistillationlarge}. However, it is still challenging to accommodate individuals with diverse evacuation patterns~\cite{zhao2022estimating, sun2024social}. For example, some individuals may evacuate immediately upon receiving an official order, prioritizing institutional guidance, while others may rely on social cues, choosing to stay until observing their neighbor evacuating.
(3) \textbf{Incorporating implicit mental states}: 
Previous studies show that integrating mental states will benefit the improvement of human behavior predictions~\cite{gu2024simpletom}. However, in a highly dynamic and chaotic wildfire environment, there exist many implicit, diverse, and complex mental states, perceptions, or beliefs, that drive wildfire evacuation behaviors, making it challenging to learn and incorporate such implicit mental states.
(4) \textbf{One mental state may map to multiple behaviors}: 
Even if individuals share the same mental state, their behaviors can vary due to external factors. For instance, two individuals with identical risk perceptions might make different evacuation decisions due to the different traffic congestion situations and shelter accessibility~\cite{collins2018evacuation, hong2020modeling}.

To address these challenges, we propose \emph{FLARE}, a novel LLM-based framework for evacuation decision prediction. We introduced risk perception and threat assessment~\cite{SUN2024106557}, two critical concepts in traditional evacuation behavioral models, to represent individual mental states. As in Figure~\ref{fig:overview}, we design a classifier based on PADM, constructed using historical datasets and empirical behavioral studies, to select the most relevant input variables to risk perception and threat assessment. We further design a reasoning pattern classifier to assign the most probable reasoning patterns. An LLM further infers the perceptions and assigns corresponding scores from the selected reasoning patterns. Finally, these perceptions — combined with external information and user inputs  — are integrated into a universal CoT template. The CoT is then fed into the LLM along with previous error records and their self-reflected rationale from the training phase for prediction. 

The proposed framework adapts empirical psychological and behavioral knowledge to inform the variable selection, CoT template construction, and model inference to constrain the over-expressiveness of LLMs on small datasets and encourage a better alignment with existing behavioral theories. Both reasoning path classifier and memory-based RL help mitigate RLHF's conflicting preference issues and tailor the prediction to individual behavior. Moreover, we calibrate implicit mental state generation chains through self-validation with a classifier based on PADM and sparsely available but implicit risk/threat-related answers in the survey data. We also augment the evacuation decision prediction with similar past error trials and their reflections from memory, guiding the model toward more accurate outcomes. We also integrate descriptive external knowledge (e.g., wildfire progression) as external cues about the individual situation, guiding the LLM to translate the inferred mental state into an accurate behavior prediction.

Our main contributions include:

\begin{itemize}
    \item We introduce a novel framework that integrates advanced reasoning capability of LLMs with psychological and behavioral theories, improving the accuracy of evacuation decision prediction in small, highly imbalanced data sets.
    \vspace{-0.75em}
    \item We design a behavioral theory-informed classifier to distinguish individual reasoning patterns, addressing conflicting preference problems while constraining the reasoning paradigm.
    \vspace{-0.75em}
    \item We introduce implicit mental state learning before predicting evacuation behaviors and augment them with external information and user input, to further improve reasoning capabilities for evacuation decision prediction.
    \vspace{-0.75em}
    \item We incorporate memory of error record and self-reflection mechanisms to refine the model’s reasoning process from mental states to behavioral predictions,  enhancing its alignment with real-world evacuation behavior.
\end{itemize}

\vspace{-0.2cm}
\section{Related Work}
\vspace{-0.2cm}

\subsection{Disaster Evacuation Decision Prediction}
\vspace{-0.2cm}
Sudden-onset natural hazards—such as wildfires, hurricanes, and earthquakes, often trigger cascading failures that result in widespread environmental damage and human displacement~\cite{xu2022seismic,li2023disasternet,wang2024scalable}. In response, critical tasks such as hazard progression tracking~\cite{chen2024soscheduler}, damage assessment~\cite{yu2024intelligent,xue2024post,li2025rapid,li2025scalable}, and large-scale evacuation become urgent priorities to mitigate impacts and minimize losses. Recent research has employed multiple methods to predict wildfire evacuation decisions.
\citet{mccaffrey2018should} employed a multinomial logistic model based on PADM, enhanced by a latent class approach, to predict various evacuation decisions in three US fire-prone counties. 
\citet{forrister2024analyzing} applied logistic and linear regression to predict risk perception, evacuation decision, and delay time.
\citet{xu2023predicting} benchmarked seven machine learning approaches (e.g., Random Forest, Classification And Regression Trees (CART), Extreme Gradient Boosting) and identified CART as the best-performing model for predicting evacuation behavior from the 2019 Kincade Fire survey.
Meanwhile, \citet{lovreglio2020calibrating} introduced the Wildfire Decision Model (WDM) calibrated via Hybrid Choice Models (HCM), incorporating latent factors like risk perception and prior experience for more accurate evacuation decision predictions.
\citet{sun4953233investigating} further integrates risk perception and threat assessment as latent variables into an HCM framework, improving prediction accuracy.
Traditional statistical models do not account for the logical flow of decision-making. HCM, in contrast, considers this process.

\vspace{-0.2cm}
\subsection{LLMs for Human Decision and Behavior Prediction}
Recent work increasingly leverages LLMs to model and predict human decision-making.
BigToM~\cite{gandhi2024understanding} evaluates LLMs’ Theory-of-Mind (ToM) capabilities using causal templates, finding that GPT-4 partially approximates human ToM reasoning, while other models lag behind.
SUVA~\cite{leng2023llm} applies probabilistic modeling to behavioral economics games, revealing that larger LLMs exhibit stronger prosocial and group-identity effects.
SUVA~\cite{leng2023llm} and ~\cite{amirizaniani2024llms} find that larger LLMs capture prosocial behavior and emotional reasoning, though still fall short of human-level comprehension.
T4D~\cite{zhou2023far} highlights LLMs' difficulty in translating inferred mental states into strategic action without structured guidance.
LELMA~\cite{mensfelt2024logic} improves reasoning reliability through symbolic consistency checks, while SimpleToM~\cite{gu2024simpletom} emphasizes the need for deliberate prompting to elicit accurate moral and behavioral judgments.
SimpleToM~\cite{gu2024simpletom} shows that while LLMs can predict mental states and behavior, they often require careful prompting to yield accurate moral or behavioral judgments. To address this, \citet{kang-etal-2023-values} propose the Value Injection Method (VIM), embedding core human values into model outputs.
However, \citet{kuribayashi-etal-2024-psychometric} argue that prompting does not inherently yield better cognitive alignment than base model probabilities.
\citet{zhu2024language} find that arithmetic-trained LLMs can outperform classical models in evaluating risky, time-delayed choices.
Still, \citet{liu2024large} point out that LLMs tend to overestimate human rationality, overlooking well-documented cognitive biases and limiting behavioral fidelity.

A recent survey paper \cite{lei2025harnessinglargelanguagemodels} reviews a growing body of work exploring the use of LLMs in disaster management, utilizing methods such as retrieval-augmented generation (RAG), instruction tuning, prompt chaining, and knowledge graph grounding. These approaches support tasks such as classifying social media content, estimating human loss~\cite{wang2024near}, generating summaries, and answering public queries, thus improving situational awareness and response coordination.
Specifically, \citet{CHEN2024104804} proposed E-KELL, a framework that grounds large language models in structured knowledge graphs constructed from disaster domain knowledge. This design improves the reliability and interpretability of disaster response decisions by embedding this rule-based knowledge into the LLM reasoning process.
\citet{yin2025crisissensellminstructionfinetunedlarge} introduced CrisisSense-LLM, which instruction-tunes LLaMA2 on a large corpus of disaster-related tweets to perform multi-label classification over event type, informativeness, and aid-related categories, thereby improving situational awareness in disaster contexts.

While these studies demonstrate the growing utility of LLMs in disaster informatics, they primarily focus on information extraction, structured representation, and retrieval. In contrast, our work centers on modeling human behavior under disaster stress by integrating behavioral theory and cognitive perception constructs into the LLM prompting process—enabling the simulation of individual decision-making pathways such as evacuation under threat.

\vspace{-0.2cm}
\section{Method}
\vspace{-0.2cm}

\begin{figure}[t]
    \centering
    \includegraphics[width=1\linewidth]{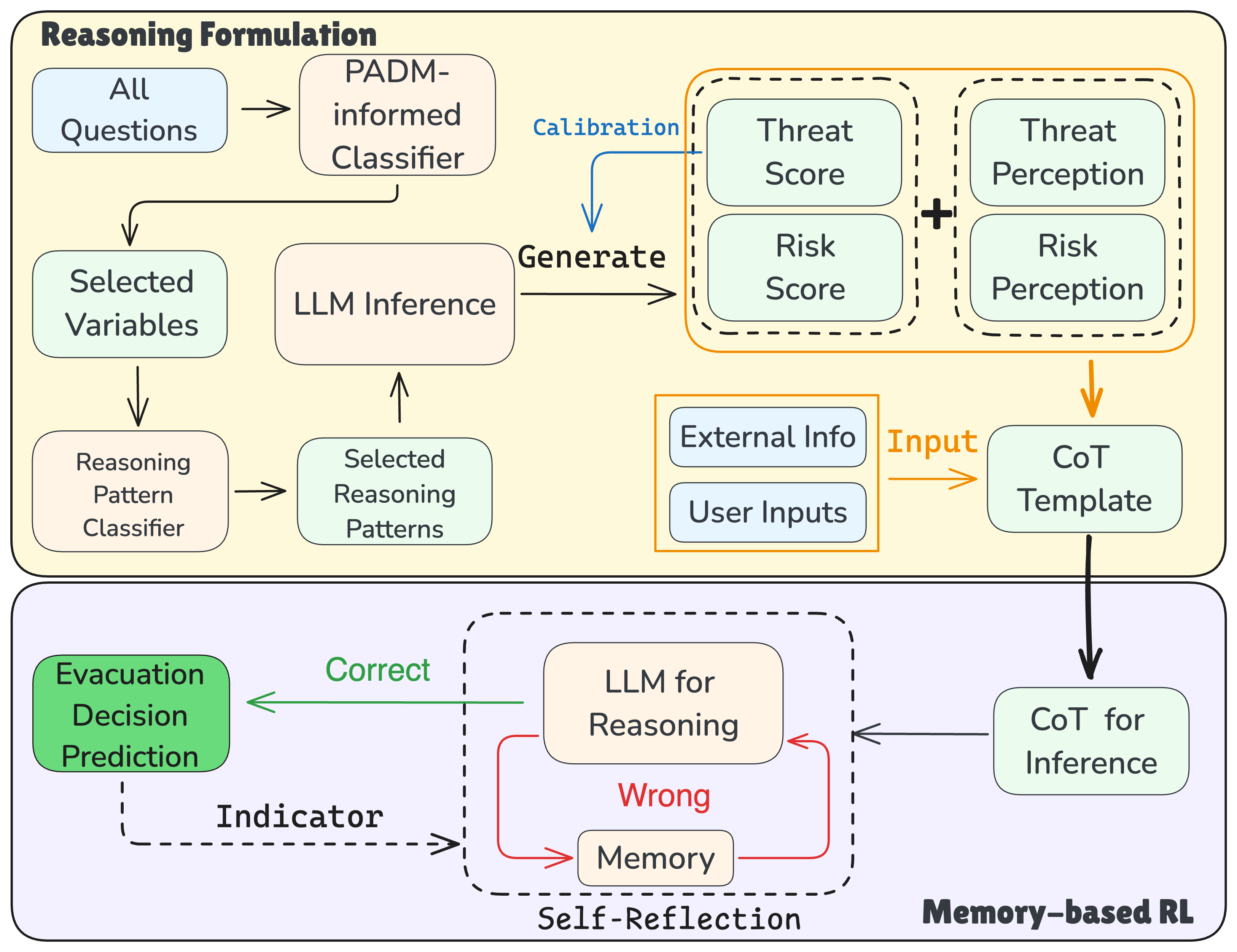}
    \caption{Overview of \emph{FLARE}.}
    \label{fig:overview}
    \vspace{-1.5em}
\end{figure}

In this section, we present the development of our LLM-based pipeline for evaluating evacuation decisions using post-wildfire survey data. The pipeline is shown in Figure \ref{fig:overview}.

\vspace{-0.2cm}
\subsection{Preliminary}
\label{sec:prel}
\vspace{-0.2cm}
\textbf{Evacuation Decision Prediction:} 
The Protective Action Decision Model (PADM) is a conceptual framework designed to explain human cognitive processes and decision-making behaviors in response to hazards and disasters. 
At its core, PADM emphasizes perceptions(e.g., threat perceptions, protective action perceptions, and stakeholder perceptions) that shape individuals’ decisions on how to respond to both immediate and long-term threats \cite{lindell2012protective}.

To advance wildfire evacuation decision prediction, researchers utilized the Hybrid Choice Model (HCM)to integrate the conceptual framework
of the PADM to predict evacuation decisions~\cite{lovreglio2020calibrating, sun4953233investigating}, quantifying two latent variables—threat assessment and risk perception — capturing individuals' assessments of wildfire threats  (e.g., injury, death) and risks (e.g., home/neighborhood damage or destruction)~\cite{Kuligowski2021}. 
By modeling how individuals process risk and threat influences, this approach provides a structured framework for understanding the cognitive processes behind wildfire evacuation behavior.

\textbf{Wildfire Evacuation Survey Data:}
The Wildfire Evacuation Survey collects data about participants' experiences during wildfire events through a set of multiple-choice, scaled and open-ended questions. The questions~\cite{kuligowski2022modeling} cover topics such as prior knowledge of wildfire risk, emergency plans, evacuation experiences, property damage, warning system awareness, and household socio-demographics. Responses range from numerical scales (e.g., perception of personal injury rated to 5) and categorical choices (e.g. Yes'' or ``No'') to counts (e.g., number of evacuations), while also capturing qualitative details like medical conditions and household composition. For detailed and complete survey refer to~\cite{KULIGOWSKI2022105541} for Kincade Fire, and ~\cite{FORRISTER2024100729} for the Marshall Fire.

\vspace{-0.2cm}

\vspace{-0.2cm}
\subsection{Reasoning Process Formulation}
\vspace{-0.2cm}

In this section, we describe how we construct classifiers based on PADM that identify the most probable reasoning patterns from survey data variables. These patterns are derived from the previously introduced risk perception and threat assessment. Once a reasoning pattern is identified, we demonstrate how an LLM generates the corresponding perception and integrates it into a CoT template, yielding the finalized CoT for evacuation prediction.

\vspace{-0.2cm}
\subsubsection{Variable Selection for Perceptions}
\vspace{-0.2cm}

Building on the HCM framework grounded in PADM for wildfire evacuation decision prediction~\cite{lovreglio2020calibrating,sun4953233investigating}, our approach aims to develop a unified statistical method to automatically select the key variables that contribute to risk perceptions and threat assessment by examining all available survey questions.

As briefly mentioned in Section \ref{sec:prel}, the survey includes questions capturing socio-demographic data, awareness and understanding, and decision-related factors such as prior wildfire risk awareness, emergency preparedness, evacuation experience, warning system awareness, personal injury perceptions, household income, employment status, and medical conditions.

In the original HCM method~\cite{sun4953233investigating}, threat and risk perceptions are validated using specific indicators derived from survey questions, requiring manual selection of variables, fitting them to indicators, and evaluating their alignment.
Our approach automates this process by incorporating all available variables into the fitting process and selecting those with the highest weights, ensuring the strongest contributions to risk perceptions and threat assessments. This minimizes bias in manual selection and enhances the model’s ability to capture key evacuation decision factors.

Formally, we regress each perception indicator on all survey variables:
\begin{equation}
Y_{k} = w_1 X_1 + w_2 X_2 + \cdots + w_n X_n + \epsilon,
\end{equation}
where $Y_k$ is the dependent variable (with $k = T_h$ for threat assessment or $k = R_h$ for risk perception), $w_i$ are the weights for the variables $X_i$, and $\epsilon$ is an error term. Once all variables are fitted in this regression, 
We then select a subset $X'$ of variables whose cumulative weight meets a predefined threshold $\theta$. Mathematically, this criterion is:
\begin{equation}
\sum_{X_i \in X'} |w_i| \ge \theta \cdot \sum_{i=1}^n |w_i|.
\end{equation}
Empirically, $\theta$ corresponds to the elbow point in the weight distribution, ensuring key variables are retained while filtering out less significant ones. The empirical results visualizing this elbow point are provided in Appendix~\ref{appx:weight_dis}, demonstrating the sharp decline in variable importance beyond the selected threshold. To justify the necessity of selecting representative variable subsets for each perception, we conducted an ablation experiment using \textit{all} survey variables without filtering. This resulted in a notable decline in performance, attributable to the transformer's limited capacity to maintain focused attention over long and densely encoded inputs. See Appendix~\ref{appx:more_result} (Table~\ref{tab:all_vars}) for further details.

Following the HCM framework, we derive four reasoning patterns by combining threat (e.g., injury, death) and risk (e.g., home/neighborhood damage or destruction) indicators-informed variable subsets, leading to four distinct reasoning processes. The selected variables are detailed in Appendix \ref{appx:select_q}.
With the core variable subsets identified, the next section explores how these guide the construction of the CoT, structuring inference pathways for evacuation decision modeling.

\vspace{-0.2cm}
\subsubsection{CoT Construction based on Perceptions}
\vspace{-0.2cm}

This section outlines our approach to constructing the Chain-of-Thought (CoT) using the variable subsets identified earlier.
We first develop a universal CoT template (Appendix~\ref{sec:Prompt}) that organizes reasoning into two behaviorally grounded stages: threat assessment followed by risk perception.
Next, we introduce a reasoning pattern classifier that selects the most likely reasoning path for each individual based on prediction success rates across candidate patterns.
Finally, we prompt the LLM to generate textual threat and risk perceptions alongside quantitative scores, establishing consistency and providing the foundation for downstream evacuation decision prediction.

\emph{\underline{Reasoning Pattern Classifier:}}
We classify individuals into the four reasoning patterns through a statistical machine learning classifier (e.g., random forest), training it on quantified survey data as input and using the LLM’s prediction performance across reasoning patterns as labels to automate pattern selection.
For each individual, we first populate all four reasoning patterns using their survey responses to generate corresponding perceptions, which are then inserted into the CoT template to form a temporary CoT for prediction. 
We then conduct multiple inference trials for each temporary CoT, with each trial producing a predicted evacuation decision (evacuate or stay) that is compared to the individual’s actual response to the evacuation decision. The success rate for each pattern is computed as the proportion of correct predictions, and the pattern with the highest success rate is considered the most probable reasoning pattern for that individual.

We further use the estimated most probable pattern as the label for individuals and train the classifier on these labels and relevant survey variables (e.g., socio-demographics, evacuation order awareness). 
This classifier automates pattern selection, ensuring that the model dynamically adapts to psychological and situational factors, enabling personalized and interpretable evacuation predictions.

\emph{\underline{Perception Inference:}} 
After selecting each individual's most probable reasoning pattern, we prompt the LLM to generate the corresponding threat perceptions and risk perceptions to construct a complete CoT for evacuation decision prediction.
The LLM first generates textual threat and risk perceptions while explicitly assigning quantitative perception scores (1–5) as calibration indicators. This dual representation enhances consistency between inferred perceptions and key survey variables. To refine this calibration, we use the first 70\% of the dataset to build a knowledge base that maps LLM-generated perceptions to survey-derived scores.

In the inference stage, we employ Retrieval-Augmented Generation (RAG) to maintain score consistency. The LLM-generated perceptions are compared to stored examples using semantic similarity, retrieving the two most similar instances (based on cosine similarity) along with their scores. 
This retrieval process aligns predicted scores with established reasoning patterns, ensuring consistency and accuracy in perception inference.

The resulting textual perceptions and calibrated scores are then integrated into the CoT template, along with contextual information extracted from survey responses (e.g., “I'm not in the area ordered to evacuate”) and user-specific inputs (Table~\ref{tab:prompt_full}), forming an individualized and perception-grounded reasoning chain for decision modeling.

\vspace{-0.2cm}
\subsection{Memory-based RL}
\vspace{-0.2cm}
Extending the previously described CoT construction for inference, we further align the reasoning process with human decision-making by incorporating an RL strategy during the LLM inference phase. 
Inspired by the verbal-based RL methodology in Reflexion~\cite{shinn2024reflexion}, our approach introduces a dedicated \emph{Memory} component that records inference errors along with the corresponding LLM-generated rationales. This \emph{Memory} mechanism enables the model to learn from past mistakes and adapt its decision-making, bridging the gap between the metal state prediction (i.e. perceptions) and evacuation decision prediction.

We begin with a training stage to construct the \emph{Memory} for subsequent use. During this stage, the actual evacuation decision reported by each respondent serves as the ground-truth reward signal. 
Whenever the LLM’s predicted decision is incorrect, we store the CoT for inference, the environment context, the LLM-generated rationale, and the correct decision in Memory. 
The model is then prompted to regenerate its reasoning and reflect on the source of the error, with these self-reflection notes also appended to \emph{Memory}.
For subsequent data samples, we retrieve the top‑$k$ most similar past entries — determined via cosine similarity over relevant variable representations — and integrate these entries as contextual information into the current inference. This retrieval mechanism allows the LLM to leverage prior cases with comparable circumstances or error patterns, refining its predictions over time.

After accumulating sufficient history in \emph{Memory} during the training phase, we transition to inference on new data. 
At this stage, self-reflection and error logging are disabled; instead, the \emph{Memory}’s contextual information is directly incorporated into the input, guiding the LLM’s reasoning process. 
The final output comprises the predicted evacuation decision and a supporting rationale derived from the CoT and contextual information retrieved from similar cases in \emph{Memory}. This comprehensive output ensures accurate predictions while providing interpretable insights into individual evacuation decisions.

\vspace{-0.2cm}
\section{Experiment}
\vspace{-0.2cm}

\emph{FLARE}, leverages both a combined dataset and the individual post-disaster survey datasets from the 2018 Carr Fire~\cite{wong2020review}, 2019 Kincade Fire~\cite{KULIGOWSKI2022105541}, and 2021 Marshall Fire~\cite{FORRISTER2024100729}. The characteristics of each dataset, including evacuation ratio, and utilized variables ratio, are detailed in Table \ref{tab:dataset_summary}. 
By integrating these data sources, we facilitate a comprehensive prediction of evacuation behavior while also preserving the unique characteristics of each event through separate analyses. The whole framework is implemented via LangChain. A detailed evaluation using metrics such as Accuracy, Precision, Recall, F1-score, Macro F1-score, and Weighted F1-score MSE is provided, with further details available in Appendix \ref{appx:implement}.

Additionally, beyond evaluating predictive accuracy, it is also important to assess the reasoning processes generated by the LLM. To this end, we compare them with open-ended questions interviewees answered in the survey (answered by 206 out of 604 interviewees), which could be treated as partial ground truth for reasoning processes. The detailed comparison can be found in Appendix \ref{appx:true_label}.

\begin{table*}[t]
\centering
\scalebox{0.7}{
\begin{tabular}{c|c|c|ccc|ccc}
\toprule
\textbf{Method} & \textbf{DataSet} & \textbf{Class} & \textbf{Precision} & \textbf{Recall} & \textbf{F1-Score} & \textbf{Accuracy} & \textbf{Macro F1} & \textbf{Weighted F1} \\
\midrule
\multirow{2}{*}{FLARE w/ GPT-4o} 
& \multirow{2}{*}{Combined Data} 
  & Stay     & 0.618 & 0.955 & 0.750 & \multirow{2}{*}{0.816} & \multirow{2}{*}{0.802} & \multirow{2}{*}{0.824} \\
&           & Evacuate & 0.976 & 0.759 & 0.823 &                     &                     &                     \\ 
\midrule
\multirow{2}{*}{FLARE w/ GPT-o3-mini} 
& \multirow{2}{*}{Combined Data} 
  & Stay     & 0.594 & 0.864 & 0.704 & \multirow{2}{*}{0.790} & \multirow{2}{*}{0.770} & \multirow{2}{*}{0.798} \\
&           & Evacuate & 0.759 & 0.837 & 0.770 &                     &                     &                     \\ 
\midrule
\multirow{2}{*}{FLARE w/ Claude-3.5}
& \multirow{2}{*}{Combined Data} 
  & Stay     & 0.850 & 0.750 & 0.810 & \multirow{2}{*}{\textbf{0.895}} & \multirow{2}{*}{\textbf{0.868}} & \multirow{2}{*}{\textbf{0.893}} \\
&           & Evacuate & 0.911 & 0.944 & 0.927 &                     &                     &                     \\ 
\midrule
\midrule
\multirow{2}{*}{Logistic Regression} 
& \multirow{2}{*}{Combined Data} 
  & Stay     & 0.560 & 0.540 & 0.550 & \multirow{2}{*}{0.697} & \multirow{2}{*}{0.640} & \multirow{2}{*}{0.679} \\
&           & Evacuate & 0.770 & 0.780 & 0.770 &                     &                     &                     \\ 
\midrule
\multirow{2}{*}{Random Forest} 
& \multirow{2}{*}{Combined Data} 
  & Stay     & 0.630 & 0.550 & 0.590 & \multirow{2}{*}{0.735} & \multirow{2}{*}{0.665} & \multirow{2}{*}{0.708} \\
&           & Evacuate & 0.780 & 0.830 & 0.860 &                     &                     &                     \\ 
\midrule
\multirow{2}{*}{GPT-4o Inference} 
& \multirow{2}{*}{Combined Data} 
  & Stay     & 0.240 & 0.310 & 0.270 & \multirow{2}{*}{0.738} & \multirow{2}{*}{0.557} & \multirow{2}{*}{0.752} \\
&           & Evacuate & 0.860 & 0.820 & 0.840 &                     &                     &                     \\ 
\midrule
\multirow{2}{*}{HCM} 
& \multirow{2}{*}{Combined Data} 
  & Stay     & 0.647 & 0.474 & 0.542 & \multirow{2}{*}{0.732} & \multirow{2}{*}{0.675} & \multirow{2}{*}{0.719} \\
&           & Evacuate & 0.761 & 0.868 & 0.809 &                     &                     &                     \\ 
\bottomrule
\end{tabular}}
\caption{\textbf{Comparison of \emph{FLARE} with baseline model on the combined dataset.} \emph{FLARE} was evaluated against four baseline methods using three different LLM backends on a combined dataset. The assessment employed metrics such as Accuracy, Macro F1, and Weighted F1, and also reported precision, recall, and F1 scores for the “Stay” and “Evacuate” classes. The results consistently demonstrate that FLARE outperforms the baseline models, regardless of the LLM employed.}
\label{tab:same_results}
\vspace{-1em}
\end{table*}

\begin{table*}[t]
\centering
\scalebox{0.7}{
\begin{tabular}{c|c|c|ccc|ccc}
\toprule
\textbf{Method} & \textbf{Train/Test Set} & \textbf{Decision} & \textbf{Precision} & \textbf{Recall} & \textbf{F1-Score} & \textbf{Accuracy} & \textbf{Macro F1} & \textbf{Weighted F1} \\
\midrule
\multirow{4}{*}{FLARE w/ Claud-3.5} 
& \multirow{2}{*}{Marshall / Kincade} 
  & Stay     & 0.433 & 0.867 & 0.578 & \multirow{2}{*}{\textbf{0.765}} & \multirow{2}{*}{\textbf{0.708}} & \multirow{2}{*}{\textbf{0.790}} \\
&           & Evacuate & 0.961 & 0.742 & 0.838 &                     &                     &                     \\ 
\cmidrule{2-9}
& \multirow{2}{*}{Kincade / Marshall} 
& Stay &  0.783 & 0.923 & 0.847 & \multirow{2}{*}{\textbf{0.870}} & \multirow{2}{*}{\textbf{0.867}} & \multirow{2}{*}{\textbf{0.871}} \\
&           & Evacuate & 0.944 & 0.836 & 0.887 &                     &                     &                     \\ 
\midrule
\multirow{4}{*}{FLARE w/ GPT-4o} 
& \multirow{2}{*}{Marshall / Kincade} 
  & Stay     & 0.387 & 0.800 & 0.522 & \multirow{2}{*}{0.728} & \multirow{2}{*}{0.668} & \multirow{2}{*}{0.757} \\
&           & Evacuate & 0.940 & 0.712 & 0.810 &                     &                     &                     \\ 
\cmidrule{2-9}
& \multirow{2}{*}{Kincade / Marshall} 
  & Stay     & 0.654 & 0.895 & 0.756 & \multirow{2}{*}{0.756} & \multirow{2}{*}{0.7556} & \multirow{2}{*}{0.756} \\
&           & Evacuate & 0.895 & 0.654 & 0.756 &                     &                     &                     \\ 
\midrule
\midrule
\multirow{4}{*}{Logistic Regression} 
& \multirow{2}{*}{Marshall / Kincade} 
  & Stay     & 0.160 & 0.190 & 0.170 & \multirow{2}{*}{0.650} & \multirow{2}{*}{0.480} & \multirow{2}{*}{0.660} \\
&           & Evacuate & 0.790 & 0.760 & 0.780 &                     &                     &                     \\ 
\cmidrule{2-9}
& \multirow{2}{*}{Kincade / Marshall} 
  & Stay     & 0.570 & 0.650 & 0.610 & \multirow{2}{*}{0.620} & \multirow{2}{*}{0.620} & \multirow{2}{*}{0.620} \\
&           & Evacuate & 0.660 & 0.590 & 0.630 &                     &                     &                     \\ 
\midrule
\multirow{4}{*}{Random Forest} 
& \multirow{2}{*}{Marshall / Kincade} 
  & Stay     & 0.000 & 0.000 & 0.000 & \multirow{2}{*}{0.800} & \multirow{2}{*}{0.450} & \multirow{2}{*}{0.720} \\
&           & Evacuate & 0.800 & 1.000 & 0.890 &                     &                     &                     \\ 
\cmidrule{2-9}
& \multirow{2}{*}{Kincade / Marshall} 
  & Stay     & 0.000 & 0.000 & 0.000 & \multirow{2}{*}{0.540} & \multirow{2}{*}{0.350} & \multirow{2}{*}{0.380} \\
&           & Evacuate & 0.540 & 1.000 & 0.700 &                     &                     &                     \\ 
\midrule
\multirow{4}{*}{GPT-4o Inference} 
& \multirow{2}{*}{Marshall / Kincade} 
  & Stay     & 0.273 & 0.231 & 0.250 & \multirow{2}{*}{0.733} & \multirow{2}{*}{0.544}& \multirow{2}{*}{0.725} \\
&           & Evacuate & 0.823 & 0.853 & 0.838 &                     &                     &                     \\ 
\cmidrule{2-9}
& \multirow{2}{*}{Kincade / Marshall} 
  & Stay     & 0.786 & 0.301 & 0.436 & \multirow{2}{*}{0.571} & \multirow{2}{*}{0.545} & \multirow{2}{*}{0.534} \\
&           & Evacuate & 0.514 & 0.900 & 0.655 &                     &                     &                     \\ 
\midrule
\multirow{4}{*}{HCM} 
& \multirow{2}{*}{Marshall / Kincade} 
  & Stay     & 0.348 & 0.736 & 0.473 & \multirow{2}{*}{0.678} & \multirow{2}{*}{0.596} & \multirow{2}{*}{0.710} \\
&           & Evacuate & 0.911 & 0.666 & 0.768 &                     &                     &                     \\ 
\cmidrule{2-9}
& \multirow{2}{*}{Kincade / Marshall} 
  & Stay     & 0.905 & 0.124 & 0.218 & \multirow{2}{*}{0.593} & \multirow{2}{*}{0.472} & \multirow{2}{*}{0.493} \\
&           & Evacuate & 0.905 & 0.989 & 0.725 &                     &                     &                     \\ 

\bottomrule
\end{tabular}}
\caption{\textbf{Comparison of \emph{FLARE} with baseline model on the cross-event dataset} derived from Kincade Fire and Marshall Fire. \emph{FLARE} was evaluated against four baseline methods using three different LLM backends on a combined dataset. The assessment employed metrics such as Accuracy, Macro F1, and Weighted F1, and also reported precision, recall, and F1 scores for the “Stay” and “Evacuate” classes. The results consistently demonstrate that FLARE has better generalizability than the baseline models}
\label{tab:cross_results}
\vspace{-1em}
\end{table*}

\vspace{-0.2cm}
\subsection{Main Results}
\vspace{-0.2cm}
We evaluate our proposed method, \emph{FLARE}, against several widely used approaches for wildfire evacuation decision prediction. We first conduct experiments on a consistent dataset to assess overall performance, followed by cross-dataset evaluations to test generalizability. 

In the consistent dataset experiments (see Table \ref{tab:same_results}), we compare the performance of our method, \emph{FLARE}—which employs three distinct backbones (GPT-o3-mini~\cite{OpenAI2025}, GPT-4o~\cite{hurst2024gpt}, and Claude-3.5~\cite{Anthropic2024}) separately—with four widely adopted prediction methods: Logistic Regression, Random Forest, LLM Inference with GPT-4o, and HCM. 
The dataset was constructed by merging survey responses from multiple wildfire events. Results on individual datasets are provided in Appendix~\ref{appx:more_result}.

The results consistently demonstrate that \emph{FLARE} achieves superior accuracy in evacuation prediction compared to the baseline methods. In contrast, the baseline approaches not only deliver lower overall accuracy but also struggle with balanced detection across various predictions, as evidenced by their F1 scores. 
Moreover, \emph{FLARE} exhibits notable adaptability across different state-of-the-art LLMs, consistently enhancing performance when employing various backbones. Notably, when using Claude-3.5 as the backbone, \emph{FLARE} improves accuracy by 13.2\%, Macro F1 by 12.7\%, and Weighted F1 by 11.9\%. These improvements indicate that advancements in LLM reasoning capabilities~\cite{Anthropic2024} could further elevate the performance of \emph{FLARE}.

In the cross-dataset generalization experiments (see Table \ref{tab:cross_results}), we use the Kincade Fire and Marshall Fire datasets in a cross-validation setup, where one dataset served as the training set and the other as the test set.
This setup is designed to account for the fact that the wildfire occurred in two different states, as illustrated in Appendix \ref{sec:geo_info}. Within the same state, evacuation laws and processes are generally similar for such events. By considering distinct states, this setup maximizes the differences between wildfire events, allowing for a more rigorous evaluation of the generalizability of our proposed methods.
As shown in Table \ref{tab:cross_results}, \emph{FLARE} achieved superior performance in terms of accuracy, Macro F1, and weighted F1 scores across both configurations. Notably, the baseline methods—particularly Logistic Regression and Random Forest—struggled to accurately classify the ``Stay'' class. In contrast, \emph{FLARE} delivered higher performance metrics and maintained a more balanced detection across classes. These results underscore the robustness of \emph{FLARE} in cross-event scenarios, highlighting its potential for effective knowledge transfer between different wildfire events.

\vspace{-0.2cm}
\subsection{Ablation Study}
\vspace{-0.2cm}

\begin{table}[htbp]
\centering
\scalebox{0.75}{
\begin{tabular}{c|ccc}
\toprule
\textbf{Method} & \textbf{Acc} & \textbf{Macro F1} & \textbf{Weighted F1} \\
\midrule
FLARE w/o CoT and RL  & 0.708  & 0.671  & 0.707 \\
FLARE w/o RL          & 0.740  & 0.727  & 0.890 \\
FLARE w/o perception & 0.768  & 0.726  & 0.756 \\
FLARE w/o CoT         & 0.773  & 0.706  & 0.768 \\
FLARE            & \textbf{0.816}  & \textbf{0.802}  & \textbf{0.824} \\
\bottomrule
\end{tabular}
}
\caption{\textbf{Ablation study of \emph{FLARE}} conducted using GPT-4 on the combined dataset. Key components of the framework were selectively removed, and the impact of each removal was evaluated using Accuracy, Macro F1, and Weighted F1 metrics. Each component's removal resulted in varying degrees of performance degradation.}
\label{tab:ablation_results}
\vspace{-1.5em}
\end{table}

We conduct ablation experiments on the combined dataset using the GPT-4o model to assess the impact of the CoT formulation and memory-based RL module on FLARE's performance. 
As shown in Table \ref{tab:ablation_results}, removing both components leads to a 13.45\% performance drop, confirming their necessity. When only the RL module is removed, the decline is less severe, highlighting the CoT formulation’s robust reasoning capability. 
Furthermore, removing only the RL module results in better performance than both the CoT and RL removed, highlighting the effectiveness of the RL module.
These findings validate that both components are essential for optimizing predictive accuracy and solidifying FLARE’s effectiveness.

\vspace{-0.2cm}
\section{Discussion - Why It Works}
\vspace{-0.2cm}

The framework's effectiveness is driven by the meticulous design of each component, enabling LLMs to generate accurate evacuation predictions through complex behavioral patterns extracted from survey data.
A key innovation in our framework is the integration of a reasoning pattern classifier with behaviorally grounded Chain-of-Thought (CoT) prompting. Prior studies show that LLMs struggle to infer human mental states without structured guidance~\cite{xu-etal-2024-opentom}. To address this, we first use a classifier to identify each individual's most likely reasoning pattern based on behavioral constructs such as threat assessment and risk perception. The selected pattern then guides the construction of a customized CoT template.
This design grounds LLM reasoning in psychological theory while adapting it to individual decision logic. The resulting CoT incorporates classifier-derived perceptions and external context, enabling the LLM to emulate complex human reasoning~\cite{wei2022chain, wang2022self, kojima2022large} and produce more accurate, interpretable evacuation predictions.

Another key component of our framework is the memory-based reinforcement learning (RL) mechanism, which integrates error correction and self-reflection to align LLM reasoning with human behavior. While LLMs can approximate mental state inference with structured CoT, their behavior prediction often degrades without detailed contextual cues~\cite{gu2024simpletom}. Yet providing excessive detail can overwhelm the model’s attention span and disrupt coherent reasoning~\cite{li2024longcontextllmsstrugglelong, levy2024tasktokensimpactinput, qian2024longllmsnecessitylongcontexttasks}.
To mitigate this, we store past inference errors along with their CoT rationales and self-generated reflections. During prediction, the model retrieves similar error cases from memory to provide targeted, context-rich cues without inflating input length. This reflective process allows the LLM to iteratively refine its reasoning, enhancing both prediction accuracy and interpretability~\cite{renze2024self, li2023reflectiontuning, li-etal-2024-selective, shinn2024reflexion}.

Additionally, our framework performs perception identification by extracting individual threat and risk perceptions through PADM-based variable selection \cite{sun4953233investigating}. This process provides structured inputs for reasoning and grounds the CoT in established behavioral theory, ensuring that the model’s inferences reflect psychologically meaningful mental states rather than generic patterns. This alignment further strengthens the connection between internal perceptions and observable evacuation behavior.

\vspace{-0.2cm}
\section{Conclusion}
\vspace{-0.2cm}

In this study, we introduced \emph{FLARE}, a novel framework that integrates the reasoning capabilities of LLMs with a well-established behavioral theory to predict the complexities of human mental states and evacuation decisions in wildfire. By systematically classifying variables and building CoT grounded in threat assessments and risk perceptions, our approach captures evacuees’ heterogeneous preferences and interprets essential perceptions. Moreover, we integrate a memory-based RL module that serves as a dynamic repository of previous errors and justifications, guiding the LLM toward spontaneously improved reasoning. Comprehensive experiments on real-world survey data from historical wildfire events demonstrate that \emph{FLARE} not only outperforms established methods but also maintains robust generalizability across different wildfire events. Overall, \emph{FLARE} introduces a novel framework to effectively integrate behavioral theory to inform and improve LLMs expressive reasoning capabilities. It enables a rigorous, theory-constrained alignment between LLM CoTs and human reasoning process, broadening the innovative use of LLMs in mimicking and predicting human behaviors.

\section*{Limitation}

Although \emph{FLARE} demonstrates promising capabilities in analyzing wildfire evacuation decisions, it is subject to several important limitations.
LLMs using CoT reasoning often lack transparency and can produce misleading outputs~\cite{turpin2023language}. This issue undermines trust and limits their adoption in policy planning and decision-making, where reliability and interpretability are essential.
Another concern is the research relies on self-reported survey data. Although the post-wildfire surveys used in this study adhere to strict data collection protocols and provide valuable insights, they were self-reported data, which may introduce potential recall bias and inaccuracies, which could affect the robustness of the conclusions.
A further issue is that, although the PADM framework accounts for geographical, meteorological, and logistical factors (e.g., perception of wildfire impact forecasts, awareness of shelter availability, and knowledge of route alternatives), our survey design did not include these elements. Consequently, our framework may not capture all factors influencing individuals' decision-making processes. Future work will incorporate these variables into the survey to facilitate more precise decision-making.
Future work should address these limitations by incorporating richer datasets that encompass a broader population and greater geographical diversity, as well as integrating more extensive environmental and logistical variables.

\section*{Acknowledgements}

This research was supported by the National Science Foundation (Award \#2442712, \#2332145, and \#2338959), and by the Rural Equitable and Accessible Transportation (REAT) Center funded by the U.S. Department of Transportation. This research was also supported by the Natural Hazards Center Quick Response Research Program, which is based on work supported by the National Science Foundation (Award \#1635593). 
Any opinions, findings, conclusions, or recommendations expressed in this material are those of the authors and do not necessarily reflect the views of NSF, USDOT, or the Natural Hazards Center.

\bibliography{custom}

\clearpage

\appendix
\section{Implementation Details}
\label{appx:implement}
\subsection{Datasets}
The survey data for this study were collected from local residents following three distinct wildfire events: the 2018 Carr Fire in California (284 responses)~\cite{wong2020review}, the 2019 Kincade Fire in California~\cite{kuligowski2022modeling}, (270 responses), and the 2021 Marshall Fire in Colorado~\cite{forrister2024analyzing}, (334 responses). Each survey covered varying aspects of evacuation behavior and perceptions. 
The Kincade Fire survey addressed pre-event and event-based factors, household characteristics, and decisions to stay or evacuate. The Marshall Fire survey emphasized pre-fire awareness, warning types, demographics, evacuation decisions and timing, and environmental cues. The Carr Fire survey captured evacuation behaviors, communication approaches, timing, transportation methods, sheltering choices, and perceived governmental response. Though similarly intended, each survey employed different questions and organizational structures. 

\begin{table*}[t]
\centering
\scalebox{0.75}{
\begin{tabular}{c|c|c|c|c|c}  
\toprule
\textbf{Dataset} & \textbf{Valid Samples} & \textbf{Fire Start Time} & \textbf{Survey Period} & \textbf{Utilized Ratio} & \textbf{Evacuation Rate}\\
\midrule
Marshall Fire   & 334 & 12/30/2021 & 5/2022 - 6/2022 & 61 / 71 & 54.19\%\\
\midrule
Kincade Fire    & 270 & 10/23/2019 & 10/2020 - 1/2021 & 66 / 77 & 81.41\% \\
\midrule
Carr Fire & 500 & 7/28/2018 & 3/2019 - 4/2019 & 71 / 75 & 89.4\% \\
\bottomrule
\end{tabular}}
\caption{Overview of wildfire evacuation dataset statistics used in our model. Valid Sample represents the number of valid survey responses. Fire Start Time indicates the date when the wildfire began. Survey Period specifies the duration over which the survey data was collected. Utilized Ratio is defined as the proportion of variables used relative to the total available variables. Evacuation Rate denotes the percentage of respondents who chose to evacuate.}
\label{tab:dataset_summary}
\end{table*}

\subsection{Implementation Details}
In this study, we combined all three wildfires' survey data into a single dataset and split it into training and test sets in an 80\%–20\% ratio for all classification models and the RL component, ensuring consistency in data usage. The classification model is a decision tree with a maximum depth set to 10, while all other hyperparameters remain at their default settings, balancing interpretability with potential model complexity. For the RL module, we similarly reserve 80\% of the data for iterative training—where the \emph{Memory} is updated repeatedly based on feedback—and use the remaining 20\% for direct inference and final performance assessment. The RL module is implemented using LangChain, providing a streamlined and reproducible framework for experimentation.

\subsection{Evaluation Metrics}
To evaluate the model's effectiveness in predicting evacuation decisions, we compare the predicted results with actual evacuation decisions using a set of well-established metrics: \textbf{Accuracy, Precision, Recall, F1-score, Macro F1-score, and Weighted F1-score}. Accuracy measures the overall correctness of predictions, while Precision and Recall assess the trade-off between false positives and false negatives, respectively. The F1 score combines Precision and Recall into a single metric to balance their trade-offs. Given the potential class imbalance in evacuation decisions, we also utilize the Macro F1-score, which averages F1-scores across all classes equally, and the Weighted F1-score, which accounts for class frequency by weighting each class's F1-score accordingly. This comprehensive multi-metric approach ensures a thorough understanding of the model's reliability and effectiveness in supporting evacuation decision-making. 

Detailed formulation of evaluation metrics shows as follow:

\textbf{Accuracy} measures the proportion of correctly classified instances among all instances and is suitable for balanced datasets. However, it may be misleading for imbalanced data. It is defined as:

\begin{equation}
Accuracy = \frac{TP + TN}{TP + TN + FP + FN}
\end{equation}

\textbf{Precision} calculates the fraction of correctly predicted positive cases out of all predicted positives. It is crucial in scenarios where false positives are costly. The formula is:

\begin{equation}
Precision = \frac{TP}{TP + FP}
\end{equation}

\textbf{Recall} measures the proportion of actual positive instances correctly identified by the model. A high recall is essential when missing positive cases is more critical than incorrectly classifying negatives. It is given by:

\begin{equation}
Recall = \frac{TP}{TP + FN}
\end{equation}

\textbf{F1-score} is the harmonic mean of precision and recall, balancing both metrics to provide a single performance measure, especially useful in imbalanced datasets. It is computed as:

\begin{equation}
F1 = 2 \times \frac{Precision \times Recall}{Precision + Recall}
\end{equation}

\textbf{Macro F1-score} computes the F1-score for each class independently and averages them, treating all classes equally. Since this is a binary classification task, it is equivalent to the standard F1-score:

\begin{equation}
Macro\ F1 = \frac{F1_{pos} + F1_{neg}}{2}
\end{equation}

\textbf{Weighted F1-score} averages F1-scores across classes but assigns a weight based on class frequency, making it more reliable for imbalanced datasets:

\begin{equation}
Weighted\ F1 = \frac{N_{pos} \times F1_{pos} + N_{neg} \times F1_{neg}}{N_{pos} + N_{neg}}
\end{equation}

where \( N_{pos} \) and \( N_{neg} \) are the number of positive and negative samples, respectively.

\textbf{Mean Squared Error (MSE)} measures the average squared difference between predicted and actual values, commonly used in regression tasks. It penalizes larger errors more heavily, making it sensitive to outliers. The formula is:

\begin{equation}
MSE = \frac{1}{n} \sum_{i=1}^{n} (y_i - \hat{y}_i)^2
\end{equation}

where \( y_i \) represents the actual value, \( \hat{y}_i \) is the predicted value, and \( n \) is the total number of samples.

\clearpage
\section{Prompt Design}
\label{sec:Prompt}

\begin{table}[htbp]
  \centering
  
  \parbox{0.98\textwidth}{
        \rule{0.98\textwidth}{1.5pt} 
        Prompt for Threat Assessment \\[-1.5mm]
        \rule{0.98\textwidth}{0.8pt} 
        
        \text{[Threat Assessment]}\\
        \textbf{System Prompt} \\
        You are an expert at rational reasoning. \\[2mm]
        \textbf{User Prompt} \\
        Analyze the following scenario: A resident is deciding whether to evacuate during a wildfire. Based on their responses to a wildfire survey, provide a brief summary of the resident's threat assessment. Response to a wildfire survey: \textit{Survey}\\[-1.5mm]
        \rule{0.98\textwidth}{1.5pt} 
}\vspace{-2mm}
\caption{\mbox{The prompt used to generate threat assessment} } 
\label{tab:prompt_threat} 
\end{table}

\vspace{-3mm}

\begin{table}[htbp]
  \centering
  \parbox{0.98\textwidth}
  {
        \rule{0.98\textwidth}{1.5pt} 
        Prompt for Risk Perception\\[-1.5mm]
        \rule{0.98\textwidth}{0.8pt} 
    
        \text{[Risk Perception]}\\
        \textbf{System Prompt} \\
        You are an expert at rational reasoning. \\[2mm]
        \textbf{User Prompt} \\
        Consider the following scenario: A resident is deciding whether to evacuate during a wildfire. Based on their Threat Assessment and their responses to a wildfire survey, briefly summarize the resident's Risk Perception. Threat Perception is: \textit{Perception}. \\
        Response to a wildfire survey: \textit{Survey}.\\[-1.5mm]
        \rule{0.98\textwidth}{0.8pt} 
    }\vspace{-2mm}
\caption{\mbox{The prompt used to generate risk perception.} } 
\label{tab:prompt_risk} 
\end{table}

\vspace{-2mm}

\begin{table}[!htbp]
  \centering
  \parbox{0.98\textwidth}{
        \rule{0.98\textwidth}{1.5pt} 
        Prompt for Evacuation Prediction \\[-1.5mm]
        \rule{0.98\textwidth}{0.8pt} 
    
        \text{[Making Decision]}\\
        \textbf{System Prompt} \\
        You are an advanced reasoning agent that can enhance your capabilities by reflecting on your own thought processes.\\[2mm]
        \textbf{User Prompt} \\
        You have access to a post-wildfire survey completed by local residents who experienced a specific wildfire event. Your task is to generate a logical, step-by-step chain of thought to infer whether the resident evacuated during the wildfire. Ensure each step is clearly connected. You must conclude with a definitive YES or NO answer regarding whether the resident evacuated. You will be provided with previous successful examples that have similar information. You may reference the rationale from these examples in your analysis.\\
        Previous Examples: \textit{Examples}\\
        Risk Perception Summary: \textit{Risk}.\\
        External information: \textit{Extras}\\[2mm]
        \textbf{Re-flexion Prompt} \\
        During the fire, this resident \textit{label} from the wildfire. Please reconsider and rethink the original questions to provide another clear and logical rationale on why the resident \textit{Label}:\\[-1.5mm]
        \rule{0.98\textwidth}{0.8pt} 

}\vspace{-2mm}
\caption{\mbox{The complete CoT used of evacuation decision prediction.} } 
\label{tab:prompt_full} 
\end{table}

\clearpage

\section{More result}
\label{appx:more_result}
\begin{table*}[t]
\centering
\scalebox{0.7}{
\begin{tabular}{c|c|c|ccc|ccc}
\toprule
\textbf{Method} & \textbf{Train/Test Set} & \textbf{Class} & \textbf{Precision} & \textbf{Recall} & \textbf{F1-Score} & \textbf{Accuracy} & \textbf{Macro F1} & \textbf{Weighted F1} \\
\midrule
\multirow{6}{*}{FLARE w/Claude-3.5} 
& \multirow{2}{*}{Kincade / Kincade} 
 & Stay & 0.765 & 0.867 & 0.813 & \multirow{2}{*}{0.926} & \multirow{2}{*}{0.883} & \multirow{2}{*}{0.928} \\
&  & Evacuate & 0.969 & 0.939 & 0.954 &                       &                       &                       \\
\cmidrule{2-9}
& \multirow{2}{*}{Marshall / Marshall} 
 & Stay & 0.875 & 0.897 & 0.886 & \multirow{2}{*}{0.910} & \multirow{2}{*}{0.906} & \multirow{2}{*}{0.910} \\
 & & Evacuate & 0.933 & 0.918 & 0.926 &                        &                        &                        \\ 
\cmidrule{2-9}
& \multirow{2}{*}{Combined Data / Combined Data} 
 & Stay & 0.850 & 0.750 & 0.810 & \multirow{2}{*}{0.895} & \multirow{2}{*}{0.868} & \multirow{2}{*}{0.893} \\
 &  & Evacuate & 0.911 & 0.944 & 0.927 &                        &                        &                        \\
\midrule
\multirow{6}{*}{FLARE w/GPT-4o} 
& \multirow{2}{*}{Kincade / Kincade} 
  & Stay     & 0.483 & 0.933 & 0.636 & \multirow{2}{*}{0.803} & \multirow{2}{*}{0.750} & \multirow{2}{*}{0.822} \\
&           & Evacuate & 0.981 & 0.773 & 0.864 &                     &                     &                     \\ 
\cmidrule{2-9}
& \multirow{2}{*}{Marshall / Marshall} 
  & Stay     & 0.729 & 0.921 & 0.814 & \multirow{2}{*}{0.822} & \multirow{2}{*}{0.821} & \multirow{2}{*}{0.823} \\
&           & Evacuate & 0.9286 & 0.750 & 0.830 &                     &                     &                     \\ 
\cmidrule{2-9}
& \multirow{2}{*}{Combined Data / Combined Data} 
  & Stay     & 0.618 & 0.955 & 0.750 & \multirow{2}{*}{0.816} & \multirow{2}{*}{0.802} & \multirow{2}{*}{0.824} \\
&           & Evacuate & 0.976 & 0.759 & 0.854 &                     &                     &                     \\ 
\midrule
\midrule
\multirow{6}{*}{Logistic Regression} 
& \multirow{2}{*}{Kincade / Kincade} 
  & Stay     & 0.600 & 0.300 & 0.400 & \multirow{2}{*}{0.780} & \multirow{2}{*}{0.630} & \multirow{2}{*}{0.746} \\
&           & Evacuate & 0.800 & 0.930 & 0.860 &                     &                     &                     \\ 
\cmidrule{2-9}
& \multirow{2}{*}{Marshall / Marshall} 
  & Stay     & 0.730 & 0.690 & 0.710 & \multirow{2}{*}{0.730} & \multirow{2}{*}{0.740} & \multirow{2}{*}{0.740} \\
&           & Evacuate & 0.730 & 0.770 & 0.750 &                     &                     &                     \\ 
\cmidrule{2-9}
& \multirow{2}{*}{Combined Data / Combined Data} 
  & Stay     & 0.560 & 0.540 & 0.550 & \multirow{2}{*}{0.697} & \multirow{2}{*}{0.640} & \multirow{2}{*}{0.679} \\
&           & Evacuate & 0.770 & 0.780 & 0.770 &                     &                     &                     \\ 
\midrule
\multirow{6}{*}{Random Forest} 
& \multirow{2}{*}{Kincade / Kincade} 
  & Stay     & 0.750 & 0.150 & 0.250 & \multirow{2}{*}{0.780} & \multirow{2}{*}{0.560} & \multirow{2}{*}{0.718} \\
&           & Evacuate & 0.780 & 0.980 & 0.870 &                     &                     &                     \\ 
\cmidrule{2-9}
& \multirow{2}{*}{Marshall / Marshall} 
  & Stay     & 0.770 & 0.710 & 0.740 & \multirow{2}{*}{0.760} & \multirow{2}{*}{0.760} & \multirow{2}{*}{0.760} \\
&           & Evacuate & 0.750 & 0.810 & 0.780 &                     &                     &                     \\ 
\cmidrule{2-9}
& \multirow{2}{*}{Combined Data / Combined Data} 
  & Stay     & 0.630 & 0.550 & 0.590 & \multirow{2}{*}{0.735} & \multirow{2}{*}{0.665} & \multirow{2}{*}{0.708} \\
&           & Evacuate & 0.780 & 0.830 & 0.860 &                     &                     &                     \\ 
\midrule
\multirow{6}{*}{LLM Inference} 
& \multirow{2}{*}{Kincade / Kincade} 
  & Stay     & 0.419 & 0.491 & 0.454 & \multirow{2}{*}{0.767} & \multirow{2}{*}{0.652} & \multirow{2}{*}{0.775} \\
&           & Evacuate & 0.870 & 0.834 & 0.852 &                     &                     &                     \\ 
\cmidrule{2-9}
& \multirow{2}{*}{Marshall / Marshall} 
  & Stay     & 0.663 & 0.810 & 0.729 & \multirow{2}{*}{0.724} & \multirow{2}{*}{0.725} & \multirow{2}{*}{0.724} \\
&           & Evacuate & 0.802 & 0.652 & 0.719 &                     &                     &                     \\ 
\cmidrule{2-9}
& \multirow{2}{*}{Combined Data / Combined Data} 
  & Stay     & 0.240 & 0.310 & 0.270 & \multirow{2}{*}{0.738} & \multirow{2}{*}{0.557} & \multirow{2}{*}{0.752} \\
&           & Evacuate & 0.860 & 0.820 & 0.840 &                     &                     &                     \\ 
\midrule
\multirow{6}{*}{HCM} 
& \multirow{2}{*}{Kincade / Kincade} 
  & Stay     & 0.633 & 0.184 & 0.244 & \multirow{2}{*}{0.811} & \multirow{2}{*}{0.481} & \multirow{2}{*}{0.761} \\
&           & Evacuate & 0.826 & 0.973 & 0.719 &                     &                     &                     \\ 
\cmidrule{2-9}
& \multirow{2}{*}{Marshall / Marshall} 
  & Stay     & 0.670 & 0.654 & 0.651 & \multirow{2}{*}{0.692} & \multirow{2}{*}{0.685} & \multirow{2}{*}{0.690} \\
&           & Evacuate & 0.714 & 0.733 & 0.719 &                     &                     &                     \\ 
\cmidrule{2-9}
& \multirow{2}{*}{Combined Data / Combined Data} 
  & Stay     & 0.647 & 0.474 & 0.542 & \multirow{2}{*}{0.732} & \multirow{2}{*}{0.675} & \multirow{2}{*}{0.719} \\
&           & Evacuate & 0.761 & 0.868 & 0.809 &                     &                     &                     \\

\bottomrule
\end{tabular}}
\caption{Performance Metrics for Different Methods and Train/Test Splits with Best Results Bolded.}
\label{tab:more_results}
\end{table*}

\subsection{Accuracy heatmaps for risk and threat perception predictions}
The accuracy  heatmaps for risk perceptions and threat assessment predictions reveals key trends in the model’s performance. Overall, the LLM demonstrates moderate accuracy, with better performance in predicting mid-range values (scores 2–4) while struggling with extreme values (scores 1 and 5). For instance, in risk perception prediction, the model performs best when the actual values are within the 2–4 range, with the highest accuracy (80\%) observed when the actual risk perception is 5, but the model predicts 3, indicating a systematic underestimation of extreme risk perceptions. Similarly, in threat assessment prediction, the model achieves its highest accuracy (50\%) when the actual threat assessment is 1, frequently predicting 2 instead. This pattern suggests that the model is biased toward moderate assessments and struggles to distinguish individuals with extremely high or low-risk perceptions or threat assessments. 

This finding suggests that while the LLM can generate reasonable approximations of threat assessment and risk perception (which are components of mental states), it struggles with capturing the extreme values that often drive actual evacuation decisions
In real-world scenarios, individuals who perceive very high risks are more likely to evacuate, whereas those with very low perceived risks may ignore warnings entirely. However, the model systematically underestimates these extremes, favoring moderate scores instead. This suggests that although it can infer general reasoning patterns, it does not fully capture the high-stakes decision-making process that translates perceptions into action. These findings align with previous research indicating that LLMs perform well in predicting human mental states but have difficulty translating those inferences into precise behavioral predictions.
The model’s tendency to underestimate extreme scores suggests the need for further calibration, like incorporating the evacuation behavioral model, reinforcement learning, and contextual variables during inferences, which is what we did in this paper.

\subsection{Further exploration in threshold settings}

To further explore justify the necessity of selecting representative variable subsets for each perception, we conducted an experiment using \textit{all} survey variables without selection. The results in Table~\ref{tab:all_vars} demonstrate a notable performance degradation compared to the original variable selection strategy. This drop is due to the transformer's limited capacity to retain and reason over long, dense input sequences—when overwhelmed with less relevant information, its attention becomes diluted and reasoning less coherent.

\subsection{Detailed Comparison with Supervised Fine-tuning methods}

As shown in Table~\ref{tab:sft_result}, the SFT model achieved moderate performance in within-dataset settings (e.g., 0.7667 accuracy on Kincade and 0.724 on Marshall). However, it struggled significantly in cross-dataset generalization, especially when training on Kincade and testing on Marshall. 
In contrast, our framework demonstrates consistent and superior performance across all configurations. These results suggest that supervised fine-tuning alone cannot effectively model the reasoning process behind evacuation behavior, whereas our structured reasoning framework grounded in behavioral theory provides better robustness and transferability.

\begin{table}[H]
\centering
\scalebox{0.75}{

\begin{tabular}{lccc}
\toprule
\textbf{Train/Test Set} & \textbf{Accuracy} & \textbf{Macro F1} & \textbf{Weighted F1} \\
\midrule
Kincade / Kincade        & 0.767  & 0.652  & 0.775  \\
Marshall / Marshall      & 0.724  & 0.725  & 0.724  \\
Marshall / Kincade       & 0.733  & 0.544  & 0.725  \\
Kincade / Marshall       & 0.6704 & 0.6703 & 0.6698 \\
Combined / Combined  & 0.738  & 0.557  & 0.752  \\
\bottomrule
\end{tabular}
}
\caption{Performance using all variables (no selection)}
\label{tab:all_vars}
\vspace{-1em}
\end{table}

\begin{table}[H]
\centering
\scalebox{0.75}{

\begin{tabular}{lccc}
\toprule
\textbf{Train/Test Set} & \textbf{Accuracy} & \textbf{Macro F1} & \textbf{Weighted F1} \\
\midrule
Kincade / Kincade        & 0.8025  & 0.7504  & 0.8222 \\
Marshall / Marshall      & 0.8222  & 0.8219  & 0.8231 \\
Marshall / Kincade       & 0.7284  & 0.6680  & 0.7569 \\
Kincade / Marshall       & 0.7556  & 0.7556  & 0.7556 \\
Combined / Combined  & 0.8158  & 0.8021  & 0.8240 \\
\bottomrule
\end{tabular}
}
\caption{Performance under original setting (thresholded variable selection)}
\label{tab:orig_vars}
\vspace{-1em}
\end{table}

\begin{table}[H]
\centering
\scalebox{0.65}{

\begin{tabular}{l l l c c c}
\toprule
\textbf{Method} & \textbf{Train/Test Set} & \textbf{Accuracy} & \textbf{Macro F1} & \textbf{Weighted F1} \\
\midrule
SFT   & Kincade / Kincade    & 0.7667 & 0.652  & 0.775  \\
SFT   & Marshall / Marshall  & 0.724  & 0.7245 & 0.724  \\
SFT   & Marshall / Kincade   & 0.733  & 0.544  & 0.725  \\
SFT   & Kincade / Marshall  & 0.5714 & 0.5453 & 0.5343 \\
SFT   & Combined / Combined  & 0.738  & 0.557  & 0.752  \\
\midrule
FLARE  & Kincade  / Kincade   & 0.803  & 0.750  & 0.822  \\
FLARE  & Marshall / Marshall  & 0.822  & 0.821  & 0.823  \\
FLARE  & Marshall / Kincade   & 0.728  & 0.668  & 0.757  \\
FLARE  & Kincade / Marshall  & 0.756  & 0.7556 & 0.756  \\
FLARE  & Combined / Combined  & 0.816  & 0.802  & 0.824  \\
\bottomrule
\end{tabular}
}
\caption{\footnotesize Comparison of Supervised Fine-Tuning (SFT) and Our Method Across Datasets}
\label{tab:sft_result}
\end{table}

\begin{figure*}[htbp]
    \centering
    \begin{subfigure}{0.48\textwidth}
        \centering
        \includegraphics[width=\linewidth]{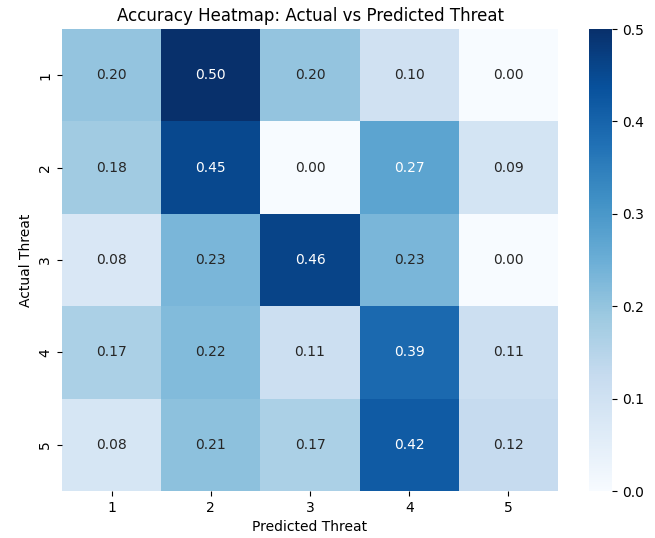}
        \caption{Accuracy map of predicted and actual threat indicator values}
        \label{fig:left}
    \end{subfigure}
    \hfill
    \begin{subfigure}{0.48\textwidth}
        \centering
        \includegraphics[width=\linewidth]{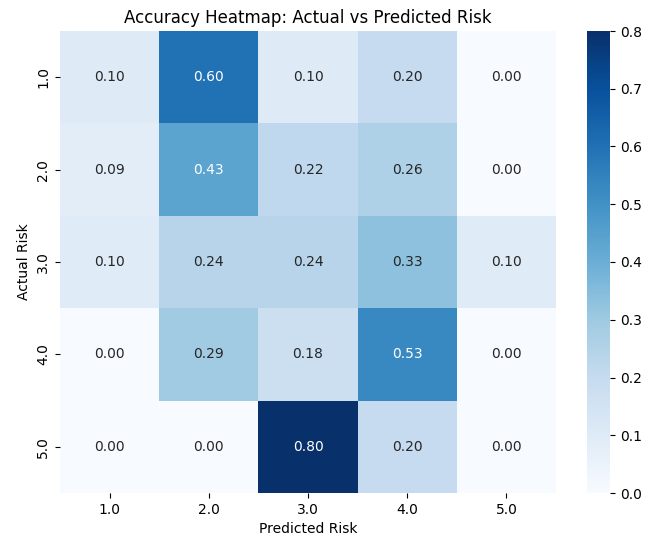}
        \caption{Accuracy map of predicted and actual risk indicator values}
        \label{fig:right}
    \end{subfigure}
    \caption{Accuracy map tested on Marshall and Kincade dataset using claude-3.5}
    \label{fig:subfig_example}
\end{figure*}

\clearpage
\clearpage

\section{Geographic Information of Wildfire Events}
\label{sec:geo_info}

\subsection{Carr Fire}

On July 23, 2018, the Carr Fire ignited in Shasta County, California, when sparks from a vehicle’s flat tire set nearby dry vegetation ablaze.

\begin{figure}[H]
    \centering
    \includegraphics[width=0.605\linewidth]{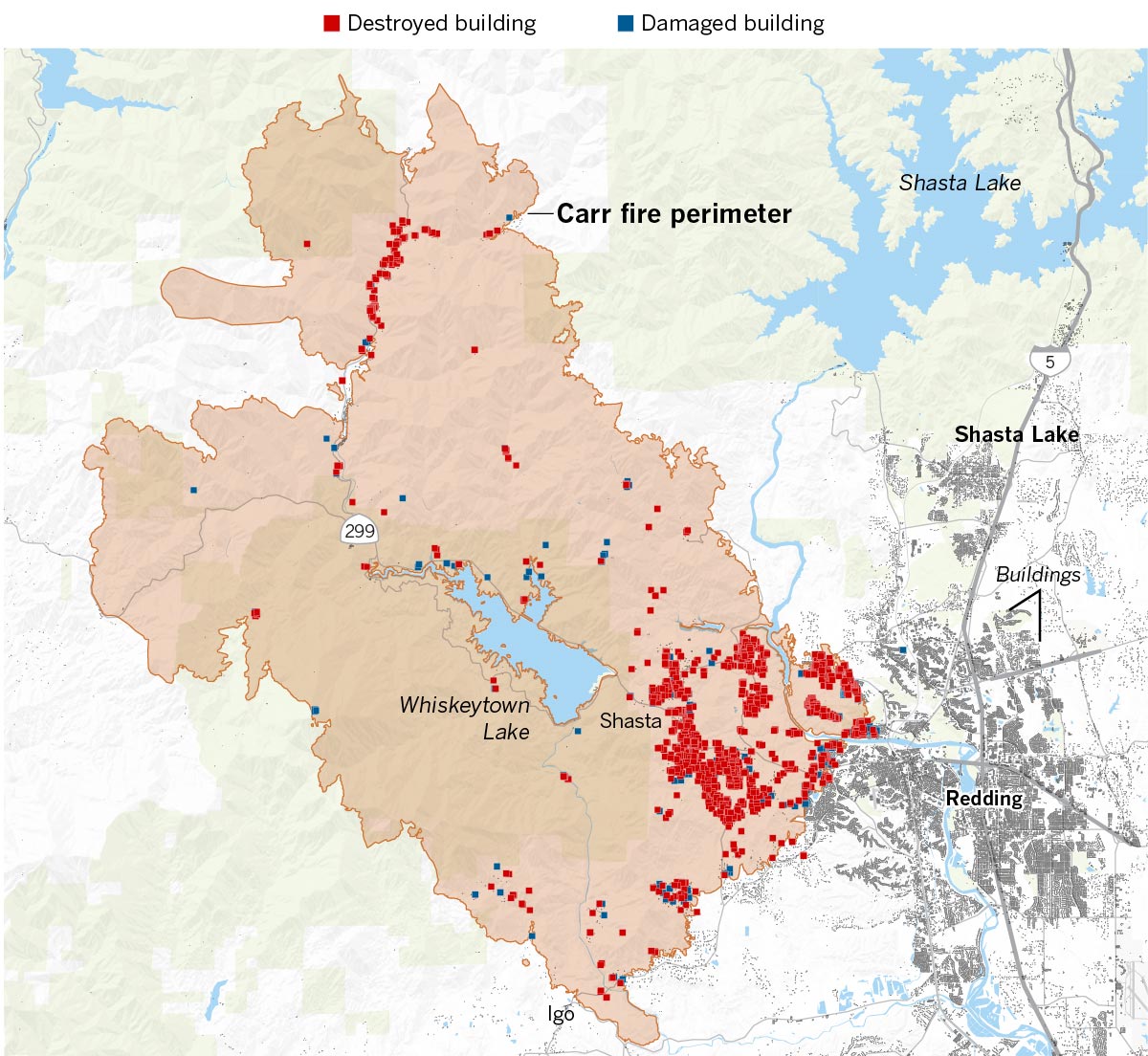}
    \caption{The damage map from the Carr Fire~\cite{schleuss2018carrfiremap}}
    \label{fig:carr_area}
\end{figure}

By August 30, 2018, the Carr Fire had been fully contained after burning over 229,000 acres, destroying approximately 1,600 structures, forcing the evacuation of around 39,000 people, claiming eight lives, and inflicting an estimated \$1.5 billion in damages. The affected area map is shown in Figure \ref{fig:carr_area}.
As it advanced rapidly to the east, the fire prompted the evacuation of French Gulch, Old Shasta, and Keswick, and worsening conditions led officials to evacuate several urban neighborhoods in Redding. Furthermore, the Carr Fire jumped the Sacramento River—partly due to fire whirls induced by the wildfire system. Ultimately, the combined efforts of 4,500 firefighting personnel and favorable weather conditions slowed its progression through Redding and surrounding rural communities, leading to its eventual containment at the end of August 2018~\cite{wong2020review}.

\subsection{Kincade Fire}
\begin{figure}[H]
    \centering
    \includegraphics[width=0.65\linewidth]{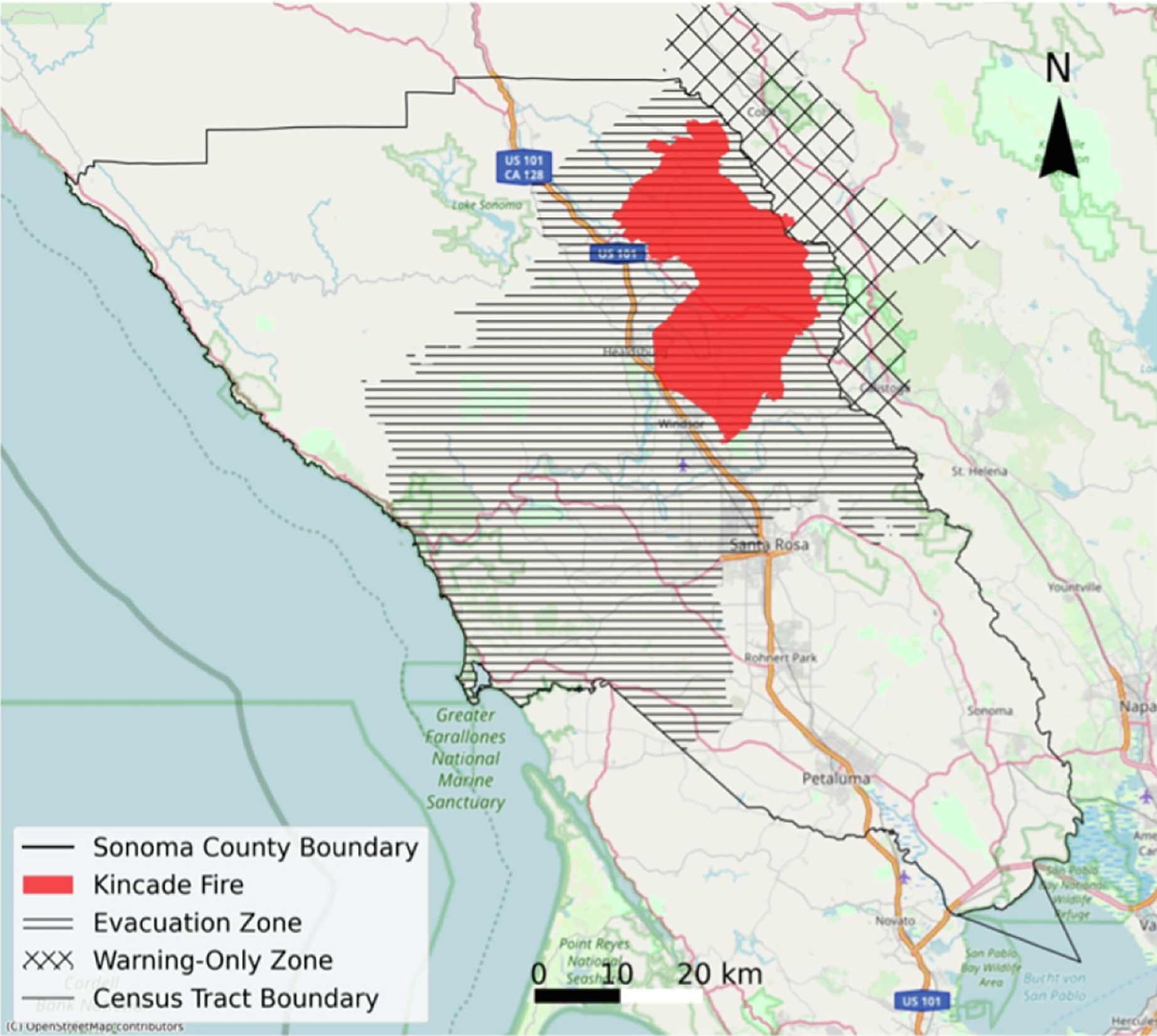}
    \caption{The wildfire impact area and evacuation area map of Kincade Fire~\cite{SUN2024106557}}
    \label{fig:kincade_area}
\end{figure}

On October 23, 2019, at 9:27 P.M., the Kincade Fire ignited northeast of Geyserville in Sonoma County, California, and was ultimately contained on November 6, 2019, at 7:00 P.M. As the largest wildfire of the 2019 California season, it burned 77,758 acres, damaged 60 structures, completely destroyed 374 structures, and injured four individuals. The event prompted the evacuation of more than 186,000 people—the largest evacuation in Sonoma County’s history. To manage this process, emergency officials partitioned the county into designated zones, issuing a mandatory evacuation order in Geyserville on October 26, followed by subsequent orders and warnings extending to areas along the Pacific Ocean and northern sections of Santa Rosa. Figure \ref{fig:kincade_area} illustrates the wildfire’s spatial impact, the delineated evacuation zones, and additional key fire parameters~\cite{SUN2024106557}.

\subsection{Marshall Fire}

The Marshall Fire ignited shortly before 10:30 a.m. on December 30, 2021, in Boulder County, Colorado, from two ignition points.

\begin{figure}[H]
    \centering
    \includegraphics[width=0.75\linewidth]{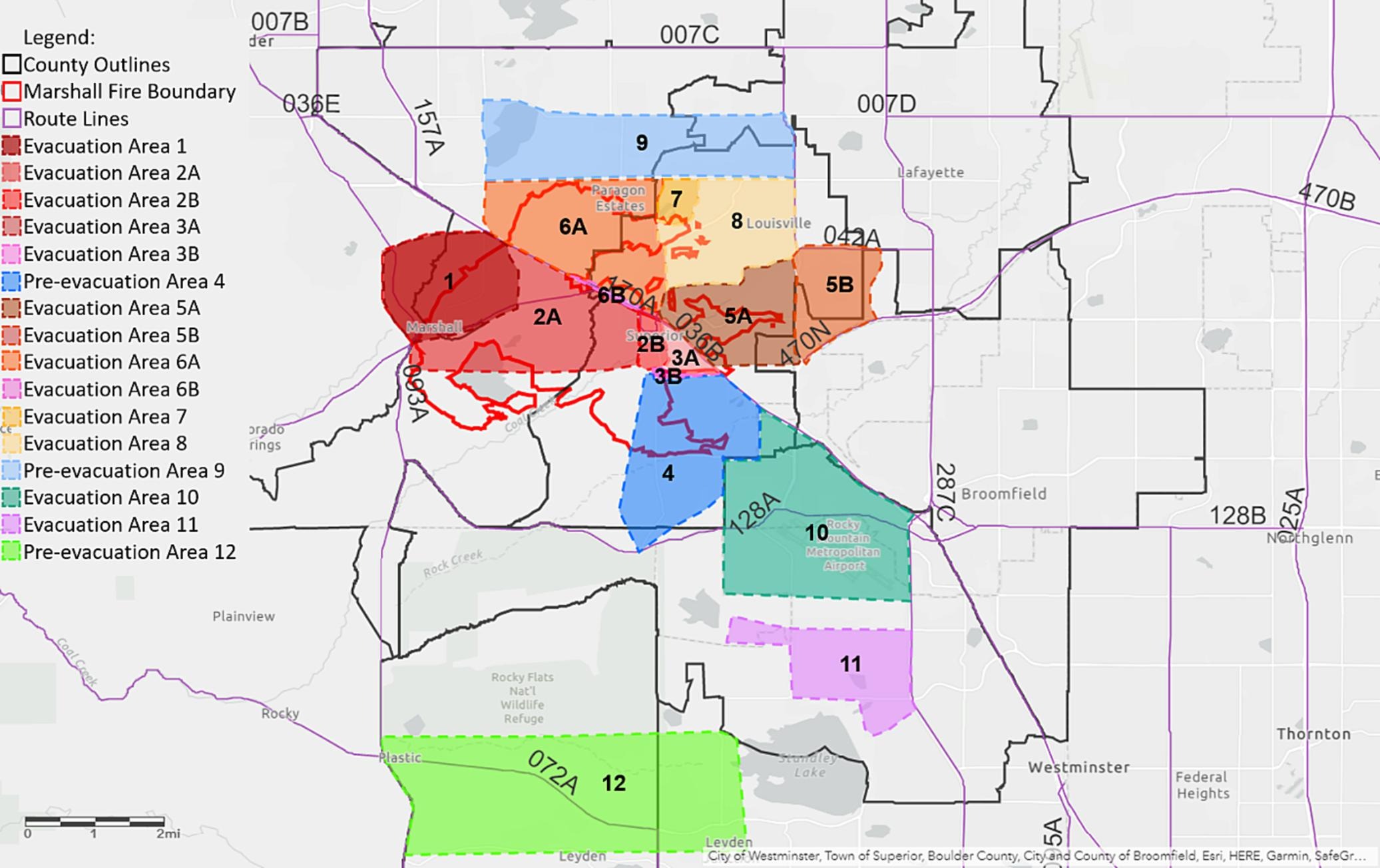}
    \caption{The wildfire impact and evacuation zones of the Marshall Fire~\cite{forrister2024analyzing}.}
    \label{fig:marshall_area}
\end{figure}

The fire quickly spread into suburban areas across Boulder, Jefferson, and Adams Counties, affecting cities such as Louisville, Superior, Broomfield, Lafayette, Arvada, and Westminster. A wet spring followed by a hot, dry summer and fall created dry fuel conditions that, combined with strong winds, accelerated the fire's spread. As Colorado’s most destructive wildfire, it burned over 6,200 acres, destroyed 1,084 homes, damaged 149 others, and caused two fatalities. In Boulder County alone, residential damages exceeded \$513 million. Over 30,000 residents were evacuated on the day of the fire. Figure~\ref{fig:marshall_area} shows the affected evacuation and pre-evacuation zones~\cite{forrister2024analyzing}.

\clearpage

\section{Weight Distribution for Different Perceptions} 
\label{appx:weight_dis}
\begin{figure*}[h]
    \centering
    \begin{subfigure}{0.45\textwidth}
        \centering
        \includegraphics[width=\linewidth]{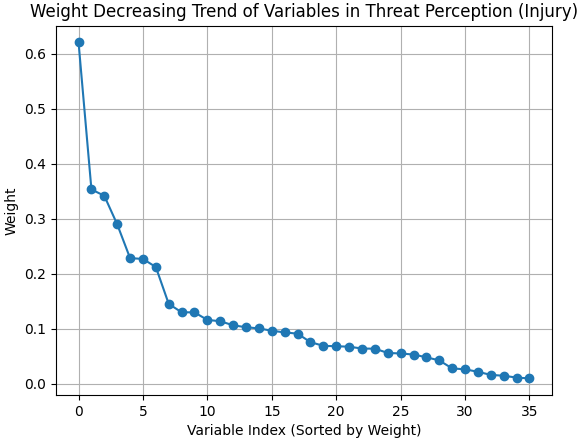}
        \caption{Weight distribution of variables for Threat Assessment (Injury)}
        \label{fig:subfig1}
    \end{subfigure}
    \hfill
    \begin{subfigure}{0.45\textwidth}
        \centering
        \includegraphics[width=\linewidth]{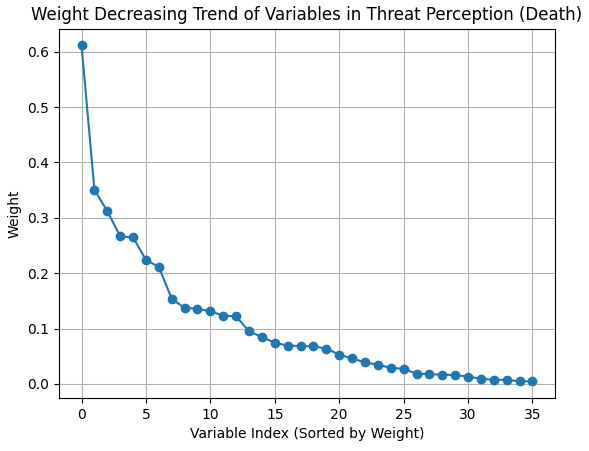}
        \caption{Weight distribution of variables for Threat Assessment (Death)}
        \label{fig:subfig2}
    \end{subfigure}
    
    \vspace{0.2cm} 
    
    \begin{subfigure}{0.45\textwidth}
        \centering
        \includegraphics[width=\linewidth]{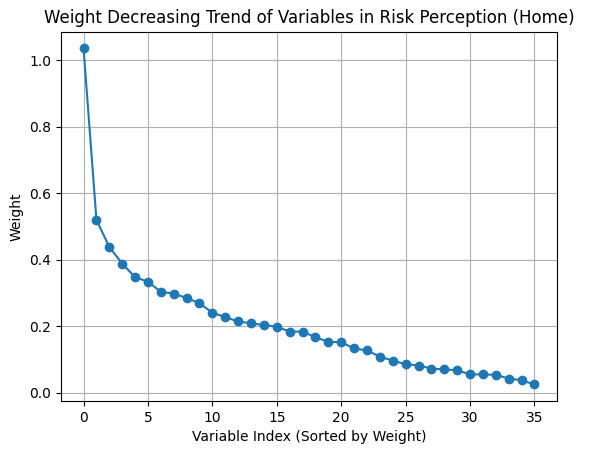}
        \caption{Weight distribution of variables for Risk Perception (Home)}
        \label{fig:subfig3}
    \end{subfigure}
    \hfill
    \begin{subfigure}{0.45\textwidth}
        \centering
        \includegraphics[width=\linewidth]{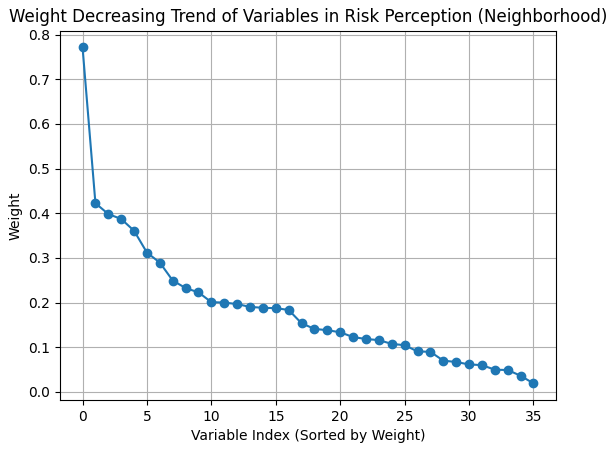}
        \caption{Weight distribution of variables for Risk Perception (Neighborhood)}
        \label{fig:subfig4}
    \end{subfigure}
    \caption{
     Examples of weight distributions from the logistic regression model are used to symmetrically select variables for each specific perception. The clear elbow points in the figure support the chosen threshold for variable selection.}
    \label{fig:main_figure}
\end{figure*}

To ensure an objective selection of key variables for each perception type, we use a logistic regression model to derive variable weights and identify an appropriate cutoff threshold. Figure \ref{fig:main_figure} illustrates the weight distribution trends for each perception category: Threat Assessment (Injury and Death) and Risk Perception (Home and Neighborhood).

Each plot displays a sharp decline in variable importance, followed by a gradual flattening, indicating the presence of an elbow point. This elbow point serves as the threshold for variable selection, ensuring that the most influential variables are retained while filtering out less significant ones.

\clearpage
\section{Selected Questions for Different Perceptions} 
\label{appx:select_q}
The selection of specific wildfire survey questions based on PADM for each threat assessment and risk perception reflects different reasoning processes in evacuation decision-making. 
An example The selected variables (i.e questions) are detailed in Table~\ref{tab:threat_questions} and Table~\ref{tab:risk_questions}

The questions in the two types of threat assessment reflect two distinct aspects. The first type is driven by direct sensory input—whether individuals observed flames or embers—and their subjective assessment of wildfire likelihood. Factors like residency duration influence familiarity with local fire risks. In contrast, the second type incorporates external cues such as warnings from social networks and educational background. Residents receiving evacuation advice from acquaintances or managing livestock may prioritize economic and logistical concerns alongside personal safety.

The questions in risk perception also follow two distinct patterns. The first emphasizes immediate physical harm, shaped by health conditions, household demographics, and emergency communications. Those with medical conditions or older adults in the household may perceive higher injury risk, while direct evacuation orders heighten urgency. The second type focuses on long-term preparedness, considering financial stability, employment, and proactive fire mitigation efforts. Residents with emergency plans or prior protective measures may perceive lower risk due to a greater sense of control.

It is also worth noting that while some variables listed in Table~\ref{tab:threat_questions} and Table~\ref{tab:risk_questions} (e.g., observing flames) may appear difficult to obtain directly during a wildfire, they can often be approximated using external sensor data such as UAV imagery or real-time fire spread models. Similarly, intermediate factors influencing decision-making can be inferred from anticipatory behavioral signals, including social media activity or traffic patterns, thereby extending FLARE’s ability to predict individual behavior even before direct wildfire exposure.

Overall, threat assessment is reactive, shaped by real-time environmental and social cues, whereas risk perception is anticipatory, centered on future consequences and preparedness. Structuring these perceptions into distinct reasoning pathways enables LLMs to model diverse decision-making profiles more effectively, improving accuracy and interpretability in wildfire evacuation predictions.

\begin{table*}[htbp]
    \centering
    \caption{Example of the Selected Survey Questions based on PADM from the survey data for Perceptions inference.}
    \label{tab:threat_questions}
    \begin{tabular}{p{0.15\linewidth} | p{0.8\linewidth}}
        \toprule
        \textbf{Category} & \textbf{Survey Question} \\
        \midrule
        \multirow{12}{*}{\makecell{Threat \\ Assessment \\ (Injury)}} 
        & 1. What was your immediate reaction to observing the flames or embers (or both)? (Select only one) \\
        & 2. Before you decided to evacuate (or stay), did you see, hear, or feel flames or embers in your immediate vicinity (that is, your neighborhood)? \\
        & 3. Before the Kincade fire, how would you have described the possibility that a wildfire would threaten your property, on a scale from 1 to 5, where 1 signifies not at all likely and 5 signifies very likely? \\
        & 4. What day did the emergency official(s) first let you know? \\
        & 5. Before you decided to evacuate (or stay), did someone you know tell you to evacuate or that a mandatory evacuation order was issued for your area? \\
        & 6. What was your immediate reaction when the emergency official(s) first let you know? (Select only one) \\
        & 7. Did you or someone in your household, including yourself, have a medical condition at the time of the Kincade fire? \\
        & 8. What time did the emergency official(s) first let you know? \\
        & 9. How long had you lived at that residence? (Select only one) \\
        & 10. Before the Kincade fire, did you know that wildfires could be a problem in your community? \\
        & 11. What were the ways people told you to evacuate or that your area was under a mandatory evacuation order? (Select all that apply) \\
        & 12. How old are you? (Please enter your age at the time you are taking this survey, below) \\
        \midrule
        \multirow{9}{*}{\makecell{Threat \\ Assessment \\ (Death)}} 
        & 1. What was your immediate reaction to observing the flames or embers (or both)? (Select only one) \\
        & 2. What was your immediate reaction when the emergency official(s) first let you know? (Select only one) \\
        & 3. Before the Kincade fire, how would you have described the possibility that a wildfire would threaten your property, on a scale from 1 to 5, where 1 signifies not at all likely and 5 signifies very likely? \\
        & 4. Before you decided to evacuate (or stay), did you see, hear, or feel flames or embers in your immediate vicinity (that is, your neighborhood)? (Mark all that apply) \\
        & 5. How old are you? (Please enter your age at the time you are taking this survey, below) \\
        & 6. Before you decided to evacuate (or stay), did someone you know tell you to evacuate or that a mandatory evacuation order was issued for your area? \\
        & 7. What is the highest level of education you have completed? (Select only one) \\
        & 8. Did you or someone in your household, including yourself, have a medical condition at the time of the Kincade fire? \\
        & 9. How many livestock or other farm animals lived in your household (or on your property) at the time the Kincade fire started (on Wednesday, October 23, 2019)? \\
        \bottomrule
    \end{tabular}
\end{table*}

\begin{table*}[htbp]
    \centering
    \caption{Selected Survey Questions based on PADM for Risk Perception}
    \label{tab:risk_questions}
    \begin{tabular}{p{0.15\linewidth} | p{0.8\linewidth}}
        \toprule
        \textbf{Category} & \textbf{Survey Question} \\
        \midrule
        \multirow{14}{*}{\makecell{Risk \\ Perception \\ (Home)}} 
        & 1. What was your immediate reaction to observing the flames or embers (or both)? (Select only one) \\
        & 2. Before you decided to evacuate (or stay), did one or more emergency officials let you know that you had to evacuate immediately and/or that your area was under a mandatory evacuation order? \\
        & 3. Before the Kincade fire, how would you have described the possibility that a wildfire would threaten your property, on a scale from 1 to 5, where 1 signifies not at all likely and 5 signifies very likely? \\
        & 4. What was your immediate reaction when the emergency official(s) first let you know? (Select only one) \\
        & 5. Did you or someone in your household, including yourself, have a medical condition at the time of the Kincade fire? \\
        & 6. If yes, what was the ultimate goal of this household emergency plan? \\
        & 7. How many adults (including your own adult children): 18 years old to 64 years old lived in your household (or on your property) at the time the Kincade fire started (on Wednesday, October 23, 2019)? \\
        & 8. How long had you lived at that residence? (Select only one) \\
        & 9. What time did the emergency official(s) first let you know? \\
        & 10. What is the highest level of education you have completed? (Select only one) \\
        & 11. Before you decided to evacuate (or stay), did you see, hear, or feel flames or embers in your immediate vicinity (that is, your neighborhood)? (Mark all that apply) \\
        & 12. Do you consider yourself (gender)? \\
        & 13. What day did the emergency official(s) first let you know? \\
        & 14. What time did you notice the fire? \\
        \midrule
        \multirow{11}{*}{\makecell{Risk \\ Perception \\ (Neighborhood)}} 
        & 1. What was your immediate reaction to observing the flames or embers (or both)? (Select only one) \\
        & 2. What was your immediate reaction when the emergency official(s) first let you know? (Select only one) \\
        & 3. Before the Kincade fire, how would you have described the possibility that a wildfire would threaten your property, on a scale from 1 to 5, where 1 signifies not at all likely and 5 signifies very likely? \\
        & 4. Before you decided to evacuate (or stay), did one or more emergency officials let you know that you had to evacuate immediately and/or that your area was under a mandatory evacuation order? \\
        & 5. If yes, what was the ultimate goal of this household emergency plan? (Select only one) \\
        & 6. What time did the emergency official(s) first let you know? \\
        & 7. How many adults (including your own adult children): 18 years old to 64 years old lived in your household (or on your property) at the time the Kincade fire started (on Wednesday, October 23, 2019)? \\
        & 8. Did you or someone in your household, including yourself, have a medical condition at the time of the Kincade fire? \\
        & 9. What day did the emergency official(s) first let you know? \\
        & 10. How long had you lived at that residence? (Select only one) \\
        \bottomrule
    \end{tabular}
\end{table*}

\clearpage

\section{Comparison with Actual Reasoning Process of Residents} 
\label{appx:true_label}

To assess the reasoning processes generated by LLMs, we compare them with open-ended responses provided by survey participants (answered by 206 out of 604 interviewees), which serve as partial ground truth for reasoning processes.

For example, the self-reported (\textbf{ground truth}) reasoning pattern of one resident is:

\begin{quote}
\itshape
“I did not believe that I was in an area at risk of wildfires, I did not think that there would be a fire, I observed others staying behind (in person OR on TV, radio, or social media), I wanted or needed to protect (or defend) my property, I wanted or needed to take care of pets or livestock (or other farm animals), I thought that it was safer to stay in place.”
\end{quote}

And the reasoning process generated by the LLM for this resident is:

\begin{quote}
\itshape
Let me create a logical chain of reasoning to determine if this resident evacuated:

Starting with pre-fire indicators: the resident had high theoretical risk awareness (5/5 rating), was well-prepared with emergency plans, and had lived in the area for over 10 years. These factors might suggest evacuation likelihood.

However, examining actual emergency response: the resident showed no reaction to official notifications, ignored mandatory evacuation orders, and explicitly planned to stay at home. They offered several reasons for staying: they did not believe the area was at risk, wanted to protect their property, and needed to care for pets or livestock.

Comparing to previous examples: Example 1 had moderate risk perception and evacuated; Example 2 had low risk perception but still evacuated. In contrast, this resident exhibited active resistance to evacuation.

Additional supporting evidence includes a large household (5 adults), multiple specific reasons for staying, and a consistent pattern of ignoring warnings.

Final analysis: the resident’s behavior demonstrates a clear and deliberate choice to stay, supported by multiple justifications and consistent dismissal of evacuation orders.

Conclusion: \textbf{NO}~--- this resident did not evacuate during the wildfire.
\end{quote}

We observe that the LLM’s reasoning converges on key motivational themes that align with the resident's own explanation. This alignment demonstrates the framework’s ability to capture nuanced behavioral logic, indicating that the generated reasoning is not only predictive but also interpretable and psychologically grounded.

\end{document}